\documentclass[10pt, letterpaper]{article}

\usepackage{times}
\usepackage{epsfig}
\usepackage{graphicx}
\usepackage{amsmath}
\usepackage{amssymb}
\usepackage{booktabs}
\usepackage[pagebackref=true,breaklinks=true,letterpaper=true,colorlinks,bookmarks=false]{hyperref}
\usepackage{flushend}
\usepackage{multirow}
\usepackage{pifont}

\begin{document}
\title{Multi-Human Parsing in the Wild}

\author{Jianshu Li $^{1*}$ \quad Jian Zhao $^{1}$\thanks{Equal contribution} \quad Yunchao Wei $^{1}$ \quad Congyan Lang $^{2}$ \\ 
\smallskip\smallskip\smallskip
Yidong Li $^{2}$ \quad Terence Sim $^{1}$ \quad Shuicheng Yan $^{1}$ \quad  Jiashi Feng$^{1}$  \\
\smallskip\smallskip\smallskip
$^{1} $National University of Singapore  \quad  $^{2}$ Beijing Jiaotong University \\
}

\maketitle

\begin{abstract}
Human parsing is attracting increasing research attention. In this work, we aim to push the frontier of human parsing by introducing the problem of multi-human parsing in the wild. Existing works on human parsing mainly tackle single-person scenarios, which deviates from real-world applications where multiple persons are present simultaneously with interaction and occlusion. To address the multi-human parsing problem, we introduce a new multi-human parsing (MHP) dataset and a novel multi-human parsing model named MH-Parser. The MHP dataset contains multiple persons captured in real-world scenes with pixel-level fine-grained semantic annotations in an instance-aware setting. The MH-Parser generates global parsing maps and person instance masks simultaneously in a bottom-up fashion with the help of a new Graph-GAN model. We envision that the MHP dataset will serve as a valuable data resource to develop new multi-human parsing models, and the MH-Parser offers a strong baseline to drive future research for multi-human parsing in the wild.
\end{abstract}

\section{Introduction}
\begin{figure*}[t]
  \centering
  \includegraphics[width=\linewidth]{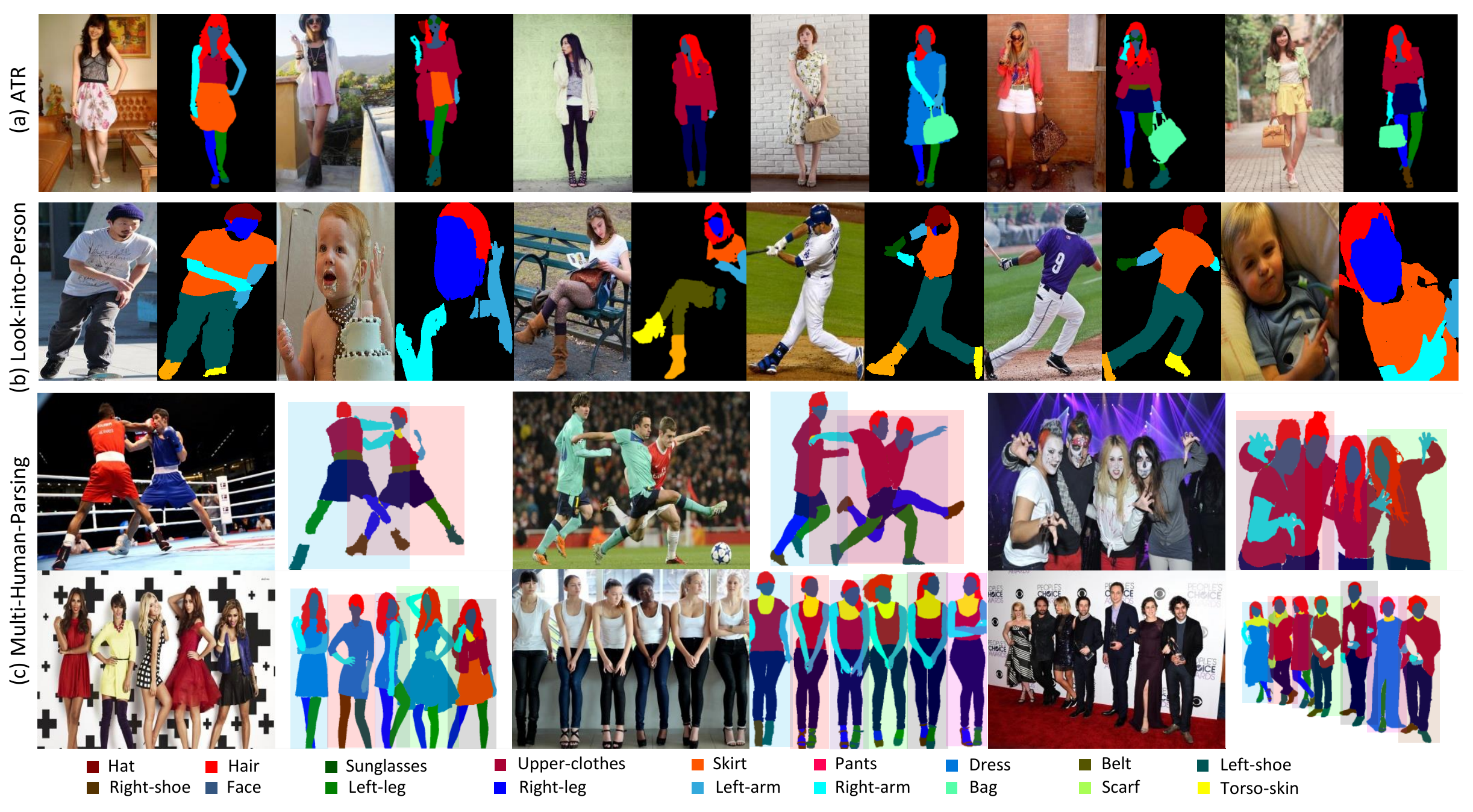}
  \caption{Annotation examples for our constructed Multiple Human Parsing (MHP) dataset (c) and other existing datasets for human parsing (a: ATR~\cite{liang2015deep}; b: Look into Person (LIP)~\cite{gong2017look}). In (c), rectangles in different colors indicate distinct person instances. ATR contains images of single persons with upright position; LIP includes more pose variations, but still only contains a single person in each image. The MHP dataset provides images with fine-grained annotations for multiple persons with interaction, occlusion and various poses, aligning better with real-world scenarios.} \label{fig:mhp_data}
\end{figure*}

Human parsing refers to partitioning persons captured in an image into multiple semantically consistent regions,~\emph{e.g.}~body parts and clothing items (cf. Fig.~\ref{fig:mhp_data}). As a fine-grained semantic segmentation task, it is more challenging than human segmentation which aims to find silhouettes of persons. Human parsing is very important for human-centric analysis and has lots of industrial applications,~\emph{e.g.} virtual reality~\cite{lin2016virtual}, video surveillance~\cite{collins2000system}, and human behavior analysis~\cite{gan2016concepts,liang2015proposal}. 

Remarkable progress has been made in parsing a single person in an image~\cite{yamaguchi2012parsing,dong2013deformable,liang2015human}. 
Single-human parsing features controlled and simplified scenarios without human interaction, occlusion or various poses that however are common in real scenarios. Thus the single-human parsing techniques deviate much from realistic requirements. Although the multi-human parsing problem can be straightforwardly solved by applying person detectors as a preprocessing step, the standard person detectors work best for upright people with simple poses, such as pedestrians. In more realistic scenarios where multiple persons are close to each other and present intimate interaction and body occlusion, person detectors tend to make false negatives, which harms the performance of multi-human parsing. Moreover, although instance semantic segmentation~\cite{liang2015proposal,dai2016instance} considers the presence of multiple humans, they only provide silhouettes of humans without fine-grained sub-category details, which does not fulfill the requirement of human parsing. Some other works~\cite{chen2014detect,chen2016deeplab,gong2017look,li2017holistic,jiang2017detangling} that look into semantic parts within persons either only consider coarse parts, or are agnostic of person instances.  

Considering the gap between current human parsing techniques and real-world requirements, we aim to drive the research on multi-human parsing.  Towards solving this challenging  problem, we introduce a new multi-human parsing dataset and a novel multi-human parsing model. In particular, we construct and annotate a new large-scale dataset,  named the Multiple Human Parsing (MHP) dataset, providing images of multiple humans in an instance-aware setting with fine-grained pixel-level  annotations. 
Humans in the images are captured in real-world scenarios with challenging poses, heavy occlusion and various appearances. Some annotation examples as well as comparison with existing human parsing datasets are illustrated in Fig.~\ref{fig:mhp_data}. See more details in Sec.~\ref{sec:mhp_data}.  The MHP dataset will serve as a valuable data resource to develop multi-human parsing models and a benchmark to evaluate their performance.

We also propose a novel  Multiple Human Parser model named MH-Parser to solve the challenging multi-human parsing problem. Unlike most existing methods focusing on single human parsing and rely on separate off-the-shelf person detectors to localize persons in images, the proposed MH-Parser tackles multiple human parsing by generating global parsing maps and instance masks for multiple persons  simultaneously in a bottom-up fashion, without resorting to any ad-hoc detection models. To better capture the human body structure, part configuration and human interaction,  the proposed MH-Parser introduces a novel Graph Generative Adversarial Network (Graph-GAN) model that learns to predict graph-structured instance parsing results by developing a graph convolutional discriminative model. The Graph-GAN is also of independent research interest for the community to apply GAN-alike models to graph data analysis.

To sum up, we make the following contributions. 1)~We introduce the multi-human parsing problem that extends the research scope of human parsing and matches real-world scenarios better. 2)~We construct the MHP dataset, a large-scale multi-human parsing benchmark, to advance the development of relevant techniques. 3)~We propose a novel model MH-Parser, which serves as a strong baseline method for multi-human parsing in the wild. 

\section{Related Work}

\paragraph{Human Parsing}
Previous human parsing methods~\cite{liu2015matching,liang2015human,liang2016semantic} and datasets~\cite{yamaguchi2012parsing,chen2014detect,liang2015human,liang2015deep} mainly focus on single-human parsing, which have severe practical limitations. None of the commonly used human parsing datasets considers instance-aware cases. Moreover, the persons in these datasets are usually in upright positions with limited pose changes, which does not accord with reality. Recently, human parsing in the wild is inspected in~\cite{gong2017look}, where persons present varying clothing appearances and diverse viewpoints, but it only considers the setting of instance-agnostic human parsing. Different from existing datasets on human parsing, the proposed MHP dataset considers simultaneous presence of multiple persons in an instance-aware setting with challenging pose variations, occlusion and interaction between persons, aligning much better with reality.

\paragraph{Instance-Aware Object/Human Segmentation}
Recently, many research efforts have been devoted to instance-aware object/human semantic segmentation. It can be solved by top-down approaches and bottom-up approaches. In the top-down family,  a detector (or a component functioning as a detector) is used to localize each instance, which is further processed to generate pixel segmentation. Multi-task Network Cascades (MNC)~\cite{dai2016instance} consists of three separate networks for differentiating instances, estimating masks and categorizing objects, receptively.  The first fully convolutional end-to-end solution to instance-aware semantic segmentation in~\cite{li2016fully} performs instance mask prediction and classification jointly. Mask-RCNN~\cite{he2017mask} adds a segmentation branch to the state-of-the-art object detector Faster-RCNN~\cite{ren2015faster} to perform instance segmentation. The top-down approaches heavily depend on the detection component, and suffer poor performance when instances are close to each other. In the bottom-up family, detection is usually not used. Usually embeddings of all pixels are learned, which are later used to cluster different pixels into different instances.  In~\cite{newell2016associative}, embeddings are learned with a grouping loss, which does pairwise comparisons across randomly sampled pixels. In~\cite{de2017semantic,neven2017fast}, a discriminative loss containing push forces and pull forces is used to learn embeddings for each pixel. In~\cite{liang2015proposal}, the embeddings of pixels are learned with direct supervision of instance locations. Different from the methods which operate on pixels, we learn an embedding of each superpixel. Furthermore, Graph-GAN is used to refine the learned embedding by leveraging high-order information. These instance-aware person segmentation methods, either top-down or bottom-up approaches, can only predict person-level segmentation without any detailed information on body parts and fashion categories, which is disadvantageous for fine-grained image understanding. In contrast, our MHP is proposed for fine-grained multi-human parsing in the wild, which aims to boost the research in real-world human-centric analysis. 

\paragraph{Generative Adversarial Networks}
The recently proposed GAN-based methods~\cite{goodfellow2014generative,radford2015unsupervised,arjovsky2017wasserstein} have yielded remarkable performance on generating photo-realistic images~\cite{huang2017beyond} and semantic segmentation maps~\cite{luc2016semantic} by specifying only a high-level goal like ``to make the output indistinguishable from the reality''~\cite{isola2016image}.  GAN automatically learns a customized loss function that  adapts to data and guides the generation process of high-quality images. Different from existing works on  image-based GANs which can only process regular input (\emph{e.g.} 2D grid images), the proposed Graph-GAN takes a flexible data structure, \emph{i.e.} graphs, as input. This is the first time that Graph-GAN was  explored in literature on GANs.

\section{The MHP Dataset}
~\label{sec:mhp_data}
In this section we introduce the Multiple Human Parsing (MHP) dataset designed for multi-human parsing in the wild. Some exemplar images  and annotations are shown in Fig.~\ref{fig:mhp_data}. 

\subsection{Image Collection and Annotation Methodology}
As pointed out in~\cite{jiang2017detangling}, in generic recognition datasets like PASCAL~\cite{everingham2015pascal} or COCO~\cite{lin2014microsoft}, only a small percentage of images contain multiple persons. Also, persons in these generic recognition datasets usually lack fine details, compared to human-centric datasets, such as those for people recognition in photo album~\cite{piper}, human immediacy prediction~\cite{chu2015multi}, interpersonal relation prediction~\cite{SOCIALRELATION_2017}, \emph{etc.} To benefit the development of new multi-human parsing models, we construct a pool of images from existing human-centric datasets~\cite{modec13,chu2015multi,SOCIALRELATION_2017,piper}, and also online Creative Commons licensed imagery. From the images pool, we select a subset of images which contain clearly visible persons with intimate interaction, rich fashion items and diverse appearances, and manually annotate them with two operations: 1)~counting and indexing the persons in the images and 2)~annotating each person. We implement an annotation tool and generate multi-scale superpixels of images based on~\cite{arbelaez2011contour} to speed up the annotation. For each instance, $18$ pre-defined semantic categories (also commonly used in  single-parsing datasets) are annotated, including \emph{hat},  \emph{hair}, \emph{sun glasses}, \emph{upper clothes}, \emph{skirt}, \emph{pants}, \emph{dress}, \emph{belt}, \emph{left shoe}, \emph{right shoe}, \emph{face}, \emph{left leg}, \emph{right leg}, \emph{left arm}, \emph{right arm}, \emph{bag}, \emph{scarf} and \emph{torso skin}. Each instance has a complete set of annotations whenever the corresponding category is present in the image. When annotating one instance, others are regarded as background. Thus, the resulting annotation set for each image consists of $P$ person-level parsing masks, where $P$ is the number of persons in the image.

\subsection{Dataset Statistics}
\begin{figure*}[t!]
    \centering
      \includegraphics[width=\linewidth]{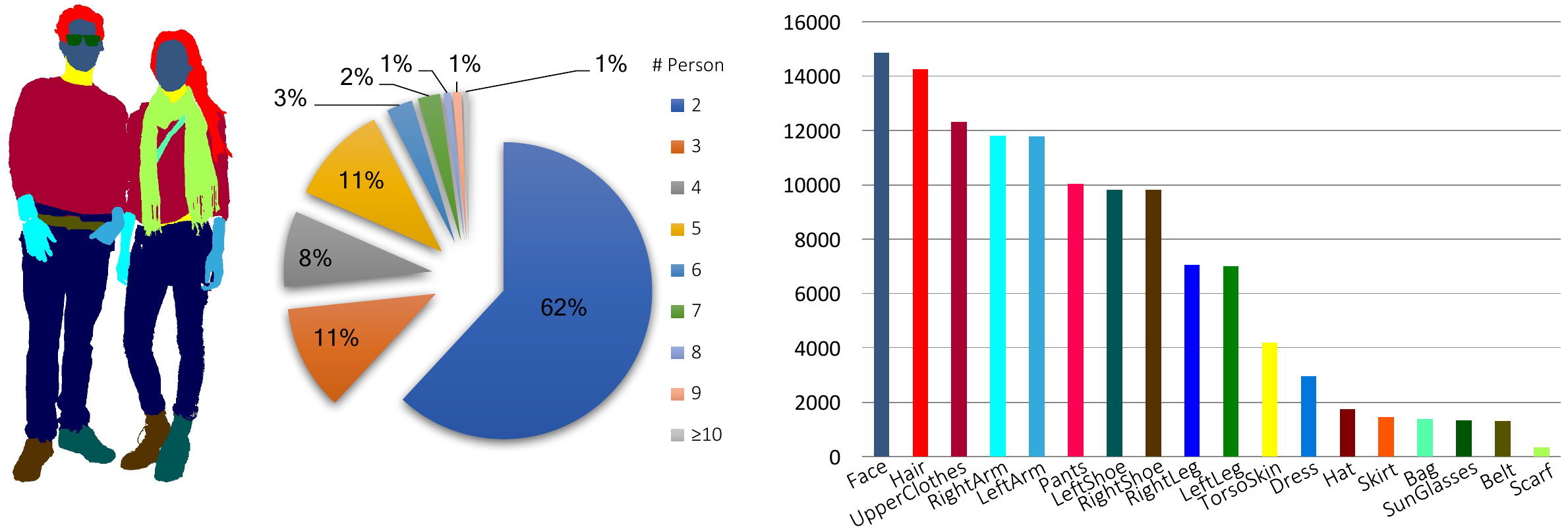} 
        \caption{Examples and statistics of the MHP dataset. Left: An annotated example for multi-human parsing. Middle: Statistics on number of persons in one image. Right: The data distribution on 18 semantic part labels in the MHP dataset.} \label{fig:num_stats}
\end{figure*}

MHP dataset contains various numbers of persons in each image, and the distribution is illustrated in Fig.~\ref{fig:num_stats} (middle). Real-world human parsing aims to analyze every detailed region of each person of interest, including different body parts, clothes and accessories. Thus we define $7$ body parts and $11$ clothing and accessory categories. Among these $7$ body parts, we divide \emph{arms} and \emph{legs} into left and right side for more precise analysis, which also increases the difficulty of the task. As for clothing categories, we have not only common clothes like \emph{upper clothes}, \emph{pants}, and \emph{shoes}, but also confusing categories such as \emph{skirt} and \emph{dress} and infrequent categories such as \emph{scarf},  \emph{sun glasses}, \emph{belt}, and \emph{bag}. The statistics for each semantic part annotation are shown in Fig.~\ref{fig:num_stats} (right).

In the MHP dataset, there are $4{,}980$ images, each with multiple persons, each with $2$-$16$ persons ($3$ on average). The resolution of the images ranges from $284\times 117$ to $6{,}919\times 4{,}511$, with an average of $755\times 734$ pixels. Totally there are $14{,}969$ person instances with fine-grained annotations at pixel-level with $18$ different  semantic labels. The resolution of each person ranges from $64\times 43$ to $2{,}627\times 3{,}881$, with an average $224\times 565$ pixels. For other human parsing datasets, Fashionista~\cite{yamaguchi2012parsing}  contains $685$ person instances, ATR~\cite{liang2015human} contains $17{,}700$ and LIP~\cite{gong2017look}  contains $50{,}462$. However, they all reflect the cases of single-human parsing, which deviates from real-world human parsing requirement. 

In MHP, the person instances are entangled with close interaction and occlusion. To verify this, we calculate the mean average Intersection Over Union (IOU) of person bounding boxes in the dataset. That is, we find the average IOU between person instances in each image, and calculate its mean value over the whole dataset. In MHP the mean average IOU is $11.71\%$. As a widely used  human instance segmentation dataset, COCO~\cite{lin2014microsoft} only has mean average IOU of $2.81\%$ for the images with multiple persons. Even for the Buffy~\cite{vineet2011human} dataset, which is used in person individuation and claims to have multiple closely entangled persons~\cite{jiang2017detangling}, the mean average IOU is only $5.93\%$. Thus  MHP is a much more challenging dataset in terms of separating closely entangled person instances. Therefore, the MHP dataset will serve as a more realistic benchmark on human-centric analysis to push the frontier of human parsing research.

\section{The MH-Parser}
\begin{figure*}[t]
  \centering
  \includegraphics[width=\linewidth]{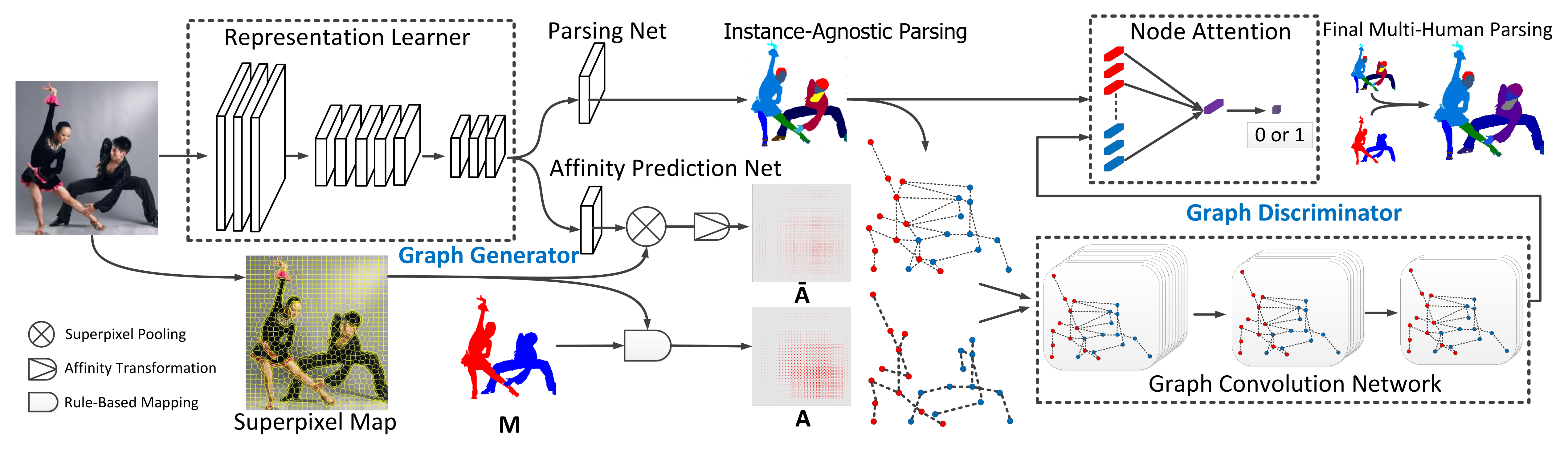}
  \caption{Architecture overview of the proposed Multiple Human Parser (MH-Parser). Here $\mathbf{M}$ refers to the global accordance map, $\mathbf{A}$ refers to the ground truth pairwise affinity map and $\bar{\mathbf{A}}$ denotes the predictions. $\mathbf{A}$ is obtained by rule-based mapping from $\mathbf{M}$ and the corresponding superpixel map (see Eqn.~\eqref{eqn:gt_pam}~and~\eqref{eqn:majority_vote}), and $\bar{\mathbf{A}}$ is the output of the graph generator (consisting of the representation learner and the affinity prediction net). The graph convolution discriminator takes the affinity graph from the graph generator as input and predicts whether it is a ground truth or a prediction. Fusing the predicted instance-agnostic parsing map and instance masks (constructed from $\bar{\mathbf{A}}$) gives the instance-aware parsing results.} \label{fig:structure}
\end{figure*}

In this section we elaborate on the proposed MH-Parser model for parsing multiple humans. The proposed MH-Parser simultaneously generates a global semantic parsing map and a pairwise affinity map (which is used to construct instance masks). The former presents  union of the instance parsing maps for all the persons in the input image, and the latter distinguishes one person from another. The overall architecture of MH-Parser is shown in Fig.~\ref{fig:structure}. 

\subsection{Global Parsing Prediction}
The MH-Parser uses a deep representation learner to learn rich and discriminative representations which are shareable for global parsing  and  affinity map prediction. In particular,  the representation learner is a fully convolutional network consisting of $101$ layers (ResNet$101$) adopted from DeepLab~\cite{chen2016deeplab}. It generates features with $1/8$ of the spatial dimension of the input image. On top of this learner, a small parsing net consisting of atrous spatial pyramid pooling~\cite{chen2016deeplab} is used to generate instance-agnostic semantic parsing maps of the whole image, as shown in Fig.~\ref{fig:structure}.

Formally, let $G_{\rm{seg}}$ denote the global parsing module. Given an input image $\mathbf{I}$ with size $H \times W$,  its output $\bar{\mathbf{S}} = G_{\rm{seg}}(\mathbf{I}) \in \mathbb{R}^{H' \times W'\times C} $ gives instance-agnostic parsing of $C$ categories with a scaled down size compared with the input image $\mathbf{I}$. The global parsing predictor $G_{\rm{seg}}$  can be trained by minimizing the following standard parsing loss:
\begin{equation}\label{eqn:seg_loss}
\small
\mathcal{L}_{\rm{seg}} (G) \triangleq\mathcal{L}_{\rm{ce}}(G_{\rm{seg}}(\mathbf{I}),\mathbf{S} ),
\end{equation}
where $\mathcal{L}_{\rm{ce}}$ is the pixel-wise cross-entropy loss and $\mathbf{S}$ is the ground truth labeling of the instance-agnostic semantic parsing map. 

\subsection{Graph-GAN for Affinity Map Prediction}
The global parsing results do not present any instance-level information which however is essential for multi-human parsing. Different from top-down solutions, we propose a novel graph-GAN model for learning instance information in a bottom-up fashion simultaneously with the global parsing prediction.

\paragraph{Global Accordance Map}
Global accordance maps distinguish different persons by associating them with different accordance scores. For an input image $\mathbf{I}$ with size $H \times W$, its global accordance map $\mathbf{M} \in \mathbb{R}^{H \times W} $ is defined as 
\begin{equation}\label{eqn:gam_r}
\small
\mathbf{M}(k)=
\begin{cases}
i,  \text{ if pixel $k$ is from the $i$-th person},  \\
0,  \text{ otherwise}. \\
\end{cases}
\end{equation}
An example of the global accordance map $\mathbf{M}$ constructed from the ground truth instance parsing map is shown in Fig.~\ref{fig:structure}. 

Predicting global accordance scores accurately is important for separating different person instances and deriving high-quality multi-human parsing results.  However, accordance prediction is very challenging, due to the large appearance variance of  intra-instance pixels and subtle difference of some pixels from different instances. The number of persons is unknown and varies for different images, making traditional classification approaches inapplicable. Moreover, the accordance scores are expected to be invariant to permutation over person instance ids. This implies that the learning process of accordance score is extremely unstable if we directly use the ground truth global accordance map defined in Eqn.~\eqref{eqn:gam_r} as supervision. 

\paragraph{Pairwise Affinity Graph}
Since directly predicting global accordance scores is difficult, MH-Parser generates a pairwise affinity graph instead. Specifically, the MH-Parser introduces a graph generator to learn to optimize the pairwise distances (or affinities) among regions within input images. In MH-Parser, superpixel is regarded as the basic unit of regions to calculate the affinities, due to the following two reasons. First, superpixels are natural low-level representations to delineate boundaries between semantic concepts. Second, superpixels can be regarded as low-level pixel grouping, so that the complexity of affinity computation is greatly reduced compared to pixel level affinity computation.

Formally, we define the pairwise affinity graph as 
\begin{equation}
 \mathcal{G}  =  (\mathcal{V}, \mathcal{E}),  \hspace{3mm}
 \begin{cases}
 v_n = s_n,  \forall n \in [1,2\cdots N], \\
e_{n_1, n_2} = \mathbf{A}(n_1, n_2).
\end{cases}
\end{equation}
In the graph, each vertex $v_n \in \mathcal{V}$  is one superpixel within the image.  There are $N$ superpixels in total and $s_n$ is the $n$-th superpixel. Each edge $e_{n_1, n_2} \in \mathcal{E}$ represents the connectivities between each pair of vertices ($v_{n_1}$, $v_{n_2}$), described by the pairwise affinity map~$\mathbf{A}$. The ground truth pairwise affinity map $\mathbf{A} \in \mathbb{R}^{N \times N}$ is derived from a rule-based mapping, which is defined as 
\begin{equation}\label{eqn:gt_pam}
\mathbf{A}(n_1,n_2)=
\begin{cases}
1,  \text{if }\sigma_{\rm{gt}}(s_{n_1}) = \sigma_{\rm{gt}}(s_{n_2})  \text{ and }  \sigma_{\rm{gt}}(s_{n_1}) >0, \\ 
0,  \text{otherwise}, \\
\end{cases}
\end{equation}
and
\begin{equation}\label{eqn:majority_vote}
\sigma_{\rm{gt}}(s_n)=\bigwedge_{k \in s_{n}} \mathbf{M}(k).
\end{equation}
Here $k \in s_n$ represents all pixels within $s_n$ and $\bigwedge$ denotes the majority vote operation. Note that although the ground truth  $\mathbf{M}$ has multiple possible values due to random assignment of person ids, the corresponding ground truth $\mathbf{A}$ is unique regardless of how the person ids are assigned. 

The pairwise affinity maps can be learned directly by taking the ground truth $\mathbf{A}$ as the regression target.  The predicted pairwise affinity map $\bar{\mathbf{A}}$ can be generated directly by an affinity prediction net, which draws features from the representation learner. The affinity prediction net first generates a set of features ${\mathbf{F}} \in \mathbb{R}^{H'' \times W'' \times C_F}$, where $C_F$ is the number of channels for $\mathbf{F}$. Then it applies superpixel pooling on $\mathbf{F}$, followed by an affinity transformation with a Gaussian kernel to obtain $\bar{\mathbf{A}} \in \mathbb{R}^{N\times N}$:
\begin{equation}\label{eqn:fake_pam}
\bar{\mathbf{A}} (n_1, n_2)= \exp\left(-\frac{\sum_{c=1}^{C_F}\left[\sigma_{\rm{sp}}(s_{n_1},c)-\sigma_{\rm{sp}}(s_{n_2,}c)\right]^2}{2\theta^2} \right),
\end{equation}
where 
\begin{equation}\label{eqn:fake_gam_ave_pooling}
\sigma_{\rm{sp}}(s_n,c)=\frac{1}{\|s_n\|}\sum\limits_{k\in s_n}\mathbf{F}(k,c).
\end{equation}
Here $\theta$ is the parameter controlling sensitivity of $\bar{\mathbf{A}}$, and $\|s_n\|$ is the number of pixels within superpixel $s_n$.

The network for predicting $\bar{\mathbf{A}}$ can be trained by minimizing the distance between $\bar{\mathbf{A}}$ and its ground truth $\mathbf{A}$. However, when learning $\bar{\mathbf{A}}$ with direct supervision, the elements within it are learned independently of each other. The contiguity and relations (reflecting intrinsic human body structures) within  $\bar{\mathbf{A}}$  are not captured. For example, if node $v_{n_1}$ is connected to $v_{n_2}$ and $v_{n_2}$ is connected to $v_{n_3}$, then $v_{n_1}$ is also connected to $v_{n_3}$. This higher-order affinity between regions is not captured for the case of direct supervision. 

\paragraph{Predicting Affinity Graph with Graph-GAN}
To remedy the potential issues in learning with direct supervision over $\mathbf{A}$, we propose a novel GAN model, Graph-GAN, to augment the learning process. Different from existing GAN-based models which can only process regular input (like 2D grid images), the Graph-GAN can take in and process flexible graph-structured data. It aims to learn high-quality affinity graphs to better capture the human body structure, part configuration and human interaction.  

In the adversarial learning of the Graph-GAN model, the ground truth  affinity graphs use $\mathbf{A}$ from Eqn.~\eqref{eqn:gt_pam} in the edge definition. The predicted  affinity graphs use  $\bar{\mathbf{A}}$ from Eqn.~\eqref{eqn:fake_pam}. The generator in Graph-GAN learns to generate high-quality affinity graphs, which are indistinguishable from the ground truth. The discriminator in Graph-GAN  targets at telling the predicted affinity graphs apart from ground truth ones. With generator and discriminator playing against each other, the discriminator learns to supervise the generator in a way tailored for the graph-structured data. 

The representation learner and the affinity prediction net are adopted as the generator in the Graph-GAN model to generate the predicted affinity graph. In order to handle graph-structured input, we propose a Graph Convolution Network (GCN) based discriminator model. The GCN~\cite{kipf2016semi,defferrard2016convolutional,manessi2017dynamic} can effectively model graph-structured data, thus is suitable for classifying input graphs and serves as the discriminator.  

In particular, we use a simple form of layer-wise propagation rule~\cite{kipf2016semi,manessi2017dynamic}:
\begin{equation}
\mathbf{H}^{(l+1)} =  f(\mathbf{H}^{(l)}, \mathbf{A}) =  \sigma(\mathbf{A}\mathbf{H}^{(l)}\mathbf{W}^{(l)}+\mathbf{b}^{(l)}),
\end{equation}
where $\mathbf{A}$ is the adjacency matrix of the graph (pairwise affinity map in our case), $\mathbf{H}$ denotes the hidden activations in GCN, $\mathbf{W}$ and $\mathbf{b}$ denote the learnable weights and biases, $\sigma$ is a non-linear activation function, and $l$ is the layer index. Thus $\mathbf{H}^{(0)}$ represents the input node features and $\mathbf{H}^{(L)}$ represents the output node features, where $L$ is the total number of layers in GCN. We follow~\cite{kipf2016semi} and normalize the adjacency matrix to make the propagation stable:

\begin{equation}\label{eqn:propagation}
\mathbf{H}^{(l+1)} =  \sigma(\hat{\mathbf{D}}^{-\frac{1}{2}}\hat{\mathbf{A}} \hat{\mathbf{D}}^{-\frac{1}{2}} \mathbf{H}^{(l)}\mathbf{W}^{(l)}+\mathbf{b}^{(l)}),
\end{equation}
where $\hat{\mathbf{A}}=\mathbf{A}+\mathbf{I}_N $ with $\mathbf{I}_N$ as the identity matrix and $\hat{\mathbf{D}}$ is the diagonal node degree matrix of $\hat{\mathbf{A}}$, \emph{i.e.} $\hat{\mathbf{D}}_{ii} = \sum_j\hat{\mathbf{A}}_{ij}$. 

In GCN, the graph convolution operation effectively diffuses the features across different regions (including body parts and background) based on the connectivities between the regions. With multiple layers of feature propagation within GCN,  higher order relations of different regions are captured, which help identify the intrinsic body part structures of multiple humans. 

Since the layer propagation rule in Eqn.~\eqref{eqn:propagation} only models the transformation of the features of nodes, node pooling operation is defined in order to obtain a graph-level feature. We define a node pooling layer on top of the final output node features with an attention mechanism, as usually used in nature language processing~\cite{lin2017structured,li2015gated}:
\begin{equation}
\mathbf{H}_g = \rm{softmax}(atten(\mathbf{H}^{(\rm{att})})) \odot \mathbf{H}^{(L)}.
\end{equation}
Here $\rm{softmax}(\rm{atten}(\mathbf{H}^{(\rm{att})}))$ generates an attention weight vector based on the attention layer input feature $\mathbf{H}^{(\rm{att})}$, and $\odot$ denotes the element-wise product, which applies the attention weight to the features of every node in the graph. We use the attention mechanism as the node feature pooling, resulting in a single descriptor $\mathbf{H}_g$ for the whole graph. With the attention pooling operation, the diffused features of different regions within an image are aggregated into one feature vector. Then $\mathbf{H}_g$ is  used as the input to a classifier to predict whether the input affinity graph is a ground truth one or a predicted one in the adversarial training setting. The input feature to the GCN model is a one-hot embedding of each node, \emph{i.e.}  $\mathbf{H}^{(0)} \in \mathbb{R}^{N\times N} =  \mathbf{I}_N$. The input feature of the node attention layer is the feature from the parsing net prediction with superpixel pooling operations applied to it, such that each node corresponds to a $C$-dimensional feature vector and $\mathbf{H}^{(\rm{att})} \in \mathbb{R}^{N\times C} $.

\subsection{Training and Inference}
We train the generator by introducing the following losses. For the global parsing task, the loss function in Eqn.~\eqref{eqn:seg_loss} is used. For the affinity graph prediction task, we minimize the distance between the predicted pairwise affinity map $\bar{\mathbf{A}}$ and the ground truth pairwise affinity map $\mathbf{A}$ with $L2$ loss:
\begin{equation}
\mathcal{L}_{L2}(G) = \|\mathbf{A} - G_{\rm{graph}}(\mathbf{I})*\mathbf{A}_{\rm{fg}} \|^2.
\end{equation}
Here $G_{\rm{graph}}(\cdot)$ represents the mapping function from the input image $\mathbf{I}$ to the predicted pairwise affinity map, \emph{i.e.} $\bar{\mathbf{A}} = G_{\rm{graph}}(\mathbf{I}) $. $\mathbf{A}_{\rm{fg}}$ is a binary mask indicating connections only between foreground nodes, and it is used to set all other connections to $0$. For training the Graph-GAN, the corresponding loss is
\begin{equation}
\mathcal{L}_{\rm{GAN}}(G,D) = \log (D(\mathbf{A})) + \log (1- D(G_{\mathrm{graph}}(\mathbf{I})*\mathbf{A}_{\rm{fg}})),
\end{equation}
where $D$ denotes the GCN-based  discriminator. Thus the overall objective function is to find $G^*$ such that
\begin{equation}\label{eqn:loss}
G^{*} = \arg \min_G \max_D   \mathcal{L}_{\rm{seg}}(G)+ \mathcal{L}_{L2}(G)+\lambda \mathcal{L}_{\mathrm{GAN}}(G,D).
\end{equation}
After finding the optimal $G^*$, we  use it to generate global parsing maps and affinity maps for testing images. 

During testing, we use the predicted affinity graph $\bar{\mathbf{A}}$ to perform spectral clustering. Background nodes are identified with the global parsing map, and are removed from the affinity graph. Then all the foreground nodes are clustered according to the pairwise affinities in $\bar{\mathbf{A}}$. Different instances of persons are identified from the clustering results. To help clustering, a regression layer built upon the representation learner is used to learn the number of persons during training, and the predicted person number is used in clustering during testing. It is omitted in the network structure and the objective function for brevity. 

\subsection{Instance Mask Refinement }
We extend our model with a refinement step to reinforce the prediction of instance masks (obtained from the clustering results) from superpixel level to pixel level. We adopt Conditional Random Field (CRF)~\cite{lafferty2001conditional} in refinement to associate each pixel in the image with one of the persons (from the clustering results) or background. The CRF model contains two  unary terms, \emph{i.e.} $ \Psi_{u} = \Psi_{\rm{Person}} +\Psi_{\rm{Global}}$ and a binary term $\Psi_{p}$. With $V_k$ denoting  the random variable for the $k$-th pixel in the image, the target of the instance mask refinement is to find the optimal solution $V_k$ for all pixels in the image that minimizes the following energy function:
\begin{equation}\label{eqn:crf_energy}
\mathit{E} = -\sum_k \ln \Psi_{u}(V_k) + \sum_{k_1<k_2} \Psi_{p}(V_{k_1}, V_{k_2}). 
\end{equation}
We define these terms as follows. Given $P$ persons from clustering results over the predicted affinity map, and assuming the $i$-th person is represented by a binary mask $\mathbf{P}^i$ indicating whether the pixel is from the $i$-th person, we define the person consistency term $\Psi_{\rm{Person}}$ as
\begin{equation}\label{eqn:person_consistency}
\Psi_{\mathrm{Person}} (V_k = i) = \mathbf{Q}_k \mathbf{P}^i_k, 
\text{ for } V_k \in [0, 1, 2, \cdots P].
\end{equation}
Here $\mathbf{Q}_k$ denotes the probability of the $k$-th pixel to be foreground. The person consistency term is designed to give strong cues about which person each foreground pixel should belong to. As in~\cite{arnab2017pixelwise,li2017holistic}, the global term $\Psi_{\rm{Global}}$ is defined as 
\begin{equation}
\Psi_{\rm{Global}} (V_k = i) = \mathbf{Q}_k,
\end{equation}
which is used to complete the person consistency term by giving equal likelihood of each foreground pixel to all the persons to correct errors in the clustering process. Finally we define our pairwise term as
\begin{equation}
\Psi_{p}(V_{k_1}, V_{k_2})= \mu(V_{k_1}, V_{k_2}) \kappa(\mathbf{f}_{k_1}, \mathbf{f}_{k_2}),
\end{equation}
where $\mu(\cdot, \cdot)$ is the compatibility function, $\kappa(\cdot, \cdot)$ is the kernel function and $\mathbf{f}_k$ is the feature vector at spatial location $k$. The feature vector contains the $C_F$-dimensional vector from $\hat{\mathbf{F}}(k)$ (obtained by up-sampling $\mathbf{F}$ to match the spatial dimension of the input image) in the affinity prediction net, the $3$-dimensional color vector $\mathbf{I}_{k}$, and the $2$-dimensional position vector $\mathbf{p}_{k}$. Thus the kernel is defined as 
\begin{equation}\label{eqn:crf_pairwise_term}
\begin{split}
\kappa(\mathbf{f}_{k_1}, \mathbf{f}_{k_2})  
& = w^{(1)} \exp\left(- \frac{\|\hat{\mathbf{F}}(k_1) -\hat{\mathbf{F}}(k_2)\|^2}{2\theta^2}\right) \\
&\hspace{1mm}+ w^{(2)} \exp\left(-\frac{\|\mathbf{p}_{k_1}-\mathbf{p}_{k_2}\|^2}{2\theta_{b_p}^2}-\frac{\|\mathbf{I}_{k_1}-\mathbf{I}_{k_2}\|^2}{2\theta_{b_I}^2}\right) \\
&\hspace{1mm}+ w^{(3)} \exp\left(-\frac{\|\mathbf{p}_{k_1}-\mathbf{p}_{k_2}\|^2}{2\theta_s^2}\right).
\end{split}
\end{equation}
In other words, the pairwise kernel consists of the learned features for pairwise distance measurement, in addition to the bilateral term and the spatial term used in~\cite{krahenbuhl2011efficient}. The compatibility function is realized by the simple Potts model.

With the above CRF model, we find the optimal solution that minimizes the energy function in Eqn.~\eqref{eqn:crf_energy} with the approximation algorithm in~\cite{krahenbuhl2011efficient} and obtain the final prediction of person instance masks for each pixel in input images. Standard CFR is also applied to the instance-agnostic parsing maps as in~\cite{chen2016deeplab}. 

\section{Experiments}
\subsection{Experimental Setup}
\paragraph{Performance Evaluation Metrics}
We use the following performance evaluation metrics for multi-human parsing. 

\emph{Average Precision based on Part (AP$^p$)}. Different from region-based Average Precision (AP$^r$) used in instance segmentation~\cite{liang2015proposal,hariharan2014simultaneous},  AP$^p$ uses part-level Intersection Over Union (IOU) of different semantic part categories within a person to determine if one instance is a true positive. Specifically, when comparing one predicted semantic part parsing map with one ground truth parsing map, we find the IOU of all the semantic part categories between them and use the average as the measure of overlap. We refer to AP under this condition as AP$^p$.  We prefer AP$^p$ over AP$^r$, as we focus on human-centric evaluation and we pay attention to how well a person as a whole is parsed. Similarly, we use AP$^p_{vol}$ to denote the average AP$^p$ values at IOU threshold from $0.1$ to $0.9$ with a step size of $0.1$. 

\emph{Percentage of Correctly Parsed Body Parts (PCP)}. As AP$^p$ averages the IOU of each part category, it cannot reflect how many parts are correctly predicted. Thus we propose to adopt PCP, originally used in human pose estimation~\cite{ferrari2008progressive,chen2014detect}, to evaluate parsing quality on the semantic parts within person instances. For each true-positive person instance, we find all the categories (excluding background) with pixel-level IOU larger than a threshold, which are regarded as correctly parsed. PCP of one person is the ratio between the correctly parsed categories and the total number of categories of that person. Missed person instances are assigned $0$ PCP. The overall PCP is the average PCP for all person instances. Note that PCP is also a human-centric evaluation metric. 

\paragraph{Datasets}
We perform experiments on the MHP dataset. From all the images in MHP,  we randomly choose $980$ images to form the testing set. The rest form a training set of $3{,}000$ images and a validation set of $1{,}000$ images. Since we are interested in the real-world situation where different people are near to each other with close interaction, we also perform experiments on the Buffy~\cite{vineet2011human} dataset as suggested in~\cite{jiang2017detangling}, which contains entangled people in almost all testing images.  

\paragraph{Implementation Details}
Due to space limit, please see supplementary material for more architecture and  implementation details.

\subsection{Experimental Analysis}

\begin{table}
  \centering
	\scriptsize
  \caption{Results from different methods on the MHP test set. The results of Mask RCNN and DL are obtained by using them to predict the instance masks, respectively, and combining with the same instance agnostic parsing map produced by MH-Parser for fair comparison. All denotes the entire test set, and Top $20\%$ and Top $5\%$ denote two subsets of testing images with top $20\%$ and top $5\%$ largest overlaps between person instances, respectively.} \label{tab:mhp_test}
  \vspace{4mm}
  \begin{tabular}{lccccccccccc}
    \toprule
\multirow{2}{*}{} & \multicolumn{3}{c}{All}	& \multicolumn{3}{c}{Top 20\%} &  \multicolumn{3}{c}{Top 5\% } \\ 
\cmidrule[\heavyrulewidth](lr){2-4}  \cmidrule[\heavyrulewidth](lr){5-7}\cmidrule[\heavyrulewidth](lr){8-10} 
& AP$^p_{0.5}$  & AP$^p_{vol}$ & PCP$_{0.5}$ &  AP$^p_{0.5}$  & AP$^p_{vol}$ & PCP$_{0.5}$ &  AP$^p_{0.5}$  & AP$^p_{vol}$ & PCP$_{0.5}$ \\
    \midrule
Detect+Parse  &29.81 & 38.83 & 43.78 & 12.08 & 30.22 &25.44 & 9.76 & 30.37 &18.36 \\
Mask RCNN~\cite{he2017mask} & \textbf{52.68} & \textbf{49.81} & \textbf{51.87} & 31.49 & 40.16 & 37.31 & 24.25 &35.63 & 28.77 \\
DL~\cite{de2017semantic} & 47.76 & 47.73 & 49.21  & 34.81 & 44.06 & 40.59 & 29.52 & 43.52 & 33.70 \\
MH-Parser &  50.10 &48.96 & 50.70  & \textbf{41.67} & \textbf{46.70} & \textbf{44.74} & \textbf{33.69} & \textbf{46.57 }& \textbf{37.01} \\
\bottomrule
  \end{tabular}
\end{table}

\subsubsection{Comparison with State-of-the-Arts}
Note that standard instance segmentation methods can only generate silhouettes of person instances and cannot produce person part parsing as desired. Thus we use them to generate instance masks as the graph generator in MH-Parser does, and combine the instance masks with the instance agnostic parsing to produce final multi-human parsing results. Here we use Mask-RCNN~\cite{he2017mask}, which is the state-of-the-art top-down model, and Discriminative Loss (DL)~\cite{de2017semantic}, a well established bottom-up model,  to generate instance masks. For Mask RCNN, we use the segmentation prediction in each detection with high confidence ($0.9$) to form the instance masks. DL can generate instance masks as the outputs of the model.  We also consider the Detect+Parse baseline method as used in traditional single human parsing, where a person detector is used to detect person instances, and a parser is used to parse each detected instance.

The performance of these methods in terms of AP$^p$, AP$^p_{vol}$ and PCP on the MHP test set is listed in Tab.~\ref{tab:mhp_test}.  In the table the overlap thresholds for AP$^p$ and PCP are both set as $0.5$. The MH-Parser, DL, the parser in Detect+Parse are trained on MHP training set with the same trunk network (ResNet$101$). Especially, DL is trained with the official code~\cite{de2017semantic,neven2017fast} with the suggested setting. The Mask-RCNN model and the detector in Detect+Parse are the top performing model with ResNet$101$ as the trunk from the official Detectron~\cite{Detectron2018}. 

We can see that the proposed MH-Parser achieves competitive performance with Mask RCNN and DL on  the MHP dataset, and outperforms Detect+Parse baseline.  To investigate how these models address the concerned challenges of closely entangled persons, we select two challenging subsets from the MHP test set. For each image in the test set, we perform a pairwise comparison of all the person instances, and find the IOU of person bounding boxes in each pair. Then the average IOU of all the pairs is used to measure the closeness of the persons in each image. One subset contains the images with top $20\%$ highest average IOUs, and the other subset contains top $5\%$. They represent images with very close interaction of human instances, reflecting the real scenarios. The results on these two subsets are listed in Tab.~\ref{tab:mhp_test}. We can see that on these challenging subsets, MH-Parse outperforms both Mask RCNN and DL. For Mask RCNN, it has difficulties to differentiate entangled persons, while as a bottom-up approach, MH-Parse can handle such cases well. For DL, it only exploits pairwise relation between embeddings of pixels, while MH-Parser models high-order relations among different regions and shows better performance. 

\paragraph{Comparison with State-of-the-Arts on Separating Person Instances}
We also evaluate the proposed MH-Parser on the Buffy dataset and compare it with other state-of-the-art methods. On Buffy forward score and backward score are used to evaluate the performance of person individuation~\cite{jiang2017detangling}. We follow the same evaluation metric, and our average forward and backward scores for Episode $4$, $5$ and $6$ on the Buffy dataset are $71.11\%$ and $71.94\%$, respectively. In~\cite{jiang2017detangling} the average forward and backward scores are $68.22\%$ and $69.66\%$ on the same dataset, and~\cite{vineet2011human} reports an average score of $62.4\%$. Note that MH-Parser is not trained on Buffy, only evaluation is performed. We can see MH-Parser achieves the best performance compared with other state-of-the-art methods in separating closely entangled persons.

\begin{table}
  \centering
    \caption{Results from different settings on the validation set. Refine refers to instance mask refinement, and Refine w/o PAM means in the refinement step the CRF is performed without the learned pairwise term from the pairwise affinity map.} \label{tab:ablation_val}
\vspace{2mm}
  \begin{tabular}{lccccccc}
    \toprule
MH-Parser 	& AP$^p_{0.5}$  &   AP$^p_{vol}$ & PCP$_{0.5}$\\
    \midrule
Baseline L2 & 41.92 & 45.21 & 46.77 \\
\hspace{2mm}+ $\mathcal{L}_{\rm{GAN}}$ & 44.34 & 46.43 & 47.62 \\
\hspace{2mm}+ Refine, w/o PAM &   49.49 & 48.98 & 50.48 \\ 
\hspace{2mm}+ Refine &   50.36 & 49.29 & 50.57 \\ 
\midrule
\hspace{2mm}w/ GT Person Number&     51.39 & 49.77 & 51.32  \\ 
\hspace{2mm}w/ GT Affinity &     55.83 & 51.28 & 55.85  \\ 
\hspace{2mm}w/ GT Global Seg. &      91.75 & 77.29 & 82.96  \\ 
    \bottomrule
  \end{tabular}
\end{table}

\subsubsection{Components Analysis for MH-Parser}

In this subsection, we test the proposed MH-Parser in various settings. All the variants of MH-Parser are trained on the MHP training set and evaluated on the validation set. The loss in Eqn.~\eqref{eqn:loss} is adjusted to either include or exclude the Graph-GAN term. We also demonstrate effects of the instance mask refinement. In the refinement, the pairwise term in Eqn.~\eqref{eqn:crf_pairwise_term} is disabled by setting $w^{(1)}$ to $0$ to investigate whether the learned pairwise term is beneficial to the refinement process. The performance of these variants in terms of AP$^p$, AP$^p_{vol}$ and PCP is listed in Tab.~\ref{tab:ablation_val}. 

From the results, we can see that compared to the $L2$ loss, the Graph-GAN can effectively improve the quality of the predicted pairwise affinity map. Better and finer affinity maps resulted from Graph-GAN help generate better grouping of the bottom level person instance information, leading to increased AP$^p$ and PCP. The instance mask refinement, especially the learned pairwise term, plays a positive role in improving the performance of multi-human parsing.

We also use the respective ground truth annotations of the three components,~\emph{i.e.} ground truth person number, ground truth affinity graph and ground truth segmentation map, to probe the upper limits of MH-Parser in Tab.~\ref{tab:ablation_val}. We can see that the person number prediction and affinity map prediction are reasonably accurate, while the global segmentation is still the major hindrance of the problem of multi-human parsing. Improvement on global segmentation can greatly boost the performance of multi-human parsing.
 
\subsubsection{Qualitative Comparison}
\begin{figure}[!t]
  \centering
  \includegraphics[width=\linewidth]{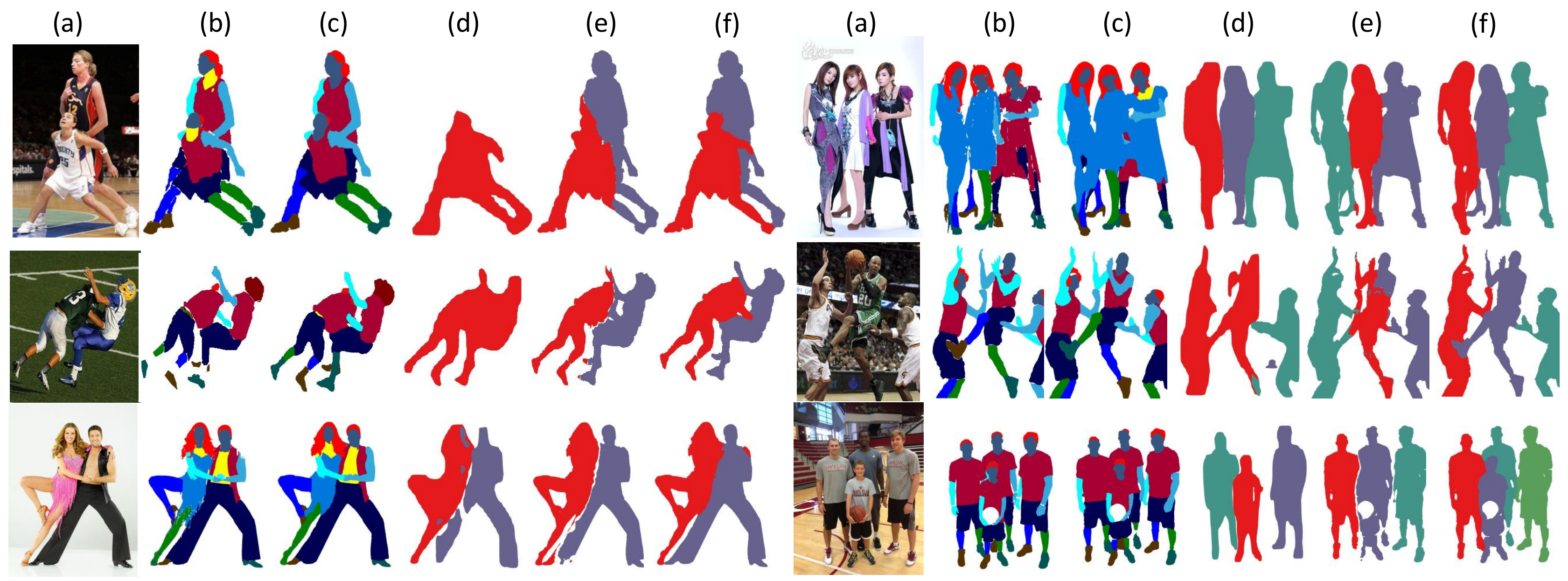}
  \caption{Visualization of parsing results. For each (a) input image, we show the (b) parsing ground truth,  (c) global parsing prediction,  person instance map predictions from (d) Mask RCNN, (e) DL and (f) MH-Parser. In (b) and (c), each color represents a semantic parsing category. In (d), (e) and (f), each color represents one person instance. We can see the proposed MH-Parser can generate satisfactory global parsing, and outperforms Mask RCNN and DL when persons are closely entangled.} \label{fig:visual}
\end{figure}

Here we visually compare the results from Mask RCNN, DL and MH-Parser. The input images, global parsing ground truths, parsing predictions, predicted instance maps from Mask RCNN, DL and MH-Parser are visualized in Fig.~\ref{fig:visual}. We can see that the MH-Parser captures both the fine-grained global parsing details and the information to differentiate person instances. For Mask RCNN, it has difficulties distinguishing closely entangled persons, especially when the bounding boxes of persons have large overlaps. The MH-Parser has better instance masks in such cases. MH-Parser also has better person instance masks than DL, especially at the boundary between two close instances. More visualized results are deferred to supplementary materials. 

\section{Conclusion}
In this paper, we tackle the multi-human parsing problem. We contributed a new large-scale MHP dataset, and also proposed a novel MH-Parser algorithm. We performed detailed evaluations of the proposed method and compared with current state-of-the-art solutions on the new benchmark dataset.  We envision that the proposed MHP dataset and the MH-Parser are promising for driving human parsing research towards real-world applications. In the future, we will make efforts to annotate a more comprehensive multiple-human parsing dataset with more images and more detailed semantic labels to further push the frontier of multiple-human parsing research.

\subsubsection*{Acknowledgments}
The work of Jianshu Li was partially funded by National Research Foundation of Singapore. The work of Jian Zhao was partially supported by National University of Defence Technology and China Scholarship Council (CSC) grant 201503170248. The work of Jiashi Feng was partially supported by National University of Singapore startup grant R-263-000-C08-133 and Ministry of Education of Singapore AcRF Tier One grant R-263-000-C21-112.

\bibliographystyle{splncs}

\end{document}